\documentclass{article}

\PassOptionsToPackage{numbers,sort}{natbib}

\usepackage[preprint]{neurips_2023}

\usepackage[utf8]{inputenc} 
\usepackage[T1]{fontenc}    
\usepackage{hyperref}       
\usepackage{url}            
\usepackage{booktabs}       
\usepackage{amsfonts}       
\usepackage{nicefrac}       
\usepackage{microtype}      
\usepackage{xcolor}         
\usepackage{graphicx}
\usepackage{pgfplots}
\usepackage{pgfplotstable}
\usepackage{booktabs}
\usepackage{makecell}
\usepackage{subcaption}

\title{From Patches to Objects: Exploiting Spatial Reasoning for Better Visual Representations}

\author{%
  Toni Albert\\
  FAU Erlangen-N\"urnberg\\
  91052 Erlangen, Germany \\
  \texttt{toni.albert@fau.de} \\
   \And
  Bjoern M.~Eskofier\\
  FAU Erlangen-N\"urnberg\\
  91052 Erlangen, Germany \\
  \texttt{bjoern.eskofier@fau.de} \\
  \And
  Dario~Zanca\\
  FAU Erlangen-N\"urnberg\\
  91052 Erlangen, Germany \\
  \texttt{dario.zanca@fau.de} \\
}

\begin{document}

\maketitle

\begin{abstract}
As the field of deep learning steadily transitions from the realm of academic research to practical application, the significance of self-supervised pretraining methods has become increasingly prominent. These methods, particularly in the image domain, offer a compelling strategy to effectively utilize the abundance of unlabeled image data, thereby enhancing downstream tasks' performance. In this paper, we propose a novel auxiliary pretraining method that is based on spatial reasoning. 
Our proposed method takes advantage of a more flexible formulation of contrastive learning by introducing spatial reasoning as an auxiliary task for discriminative self-supervised methods. 
Spatial Reasoning works by having the network predict the relative distances between sampled non-overlapping patches. We argue that this forces the network to learn more detailed and intricate internal representations of the objects and the relationships between their constituting parts. Our experiments demonstrate substantial improvement in downstream performance in linear evaluation compared to similar work and provide directions for further research into spatial reasoning.
\end{abstract}

\section{Introduction}

The rapid growth of deep learning models has led to state-of-the-art architectures containing millions of parameters \cite{Han.2022}. Concurrently, the increasing availability of unlabeled data necessitates efficient methods for reducing manual annotation. Self-supervised pretraining enables models to understand relevant domains before finetuning on a smaller labeled dataset \cite{JMLR:v11:erhan10a,Huh.30082016}.

Two primary self-supervised learning approaches exist for images \cite{liu2021self}: generative and discriminative. Generative methods reconstruct missing image components, yielding superior performance at the cost of larger networks and greater data demands. The complexity of the generative task requires the adoption of large models, such as ViT-B/16, comprising approximately 86 million parameters \cite{Dosovitskiy.22102020}. In contrast, discriminative self-supervised methods emphasize learning to differentiate between various features or patterns within the data without relying on explicit class labels. Discriminative approaches are successfully employed for pretraining of significantly smaller models like ResNet-32 \cite{he2015deep}, containing around 0.5 million parameters, thereby promoting suitability for smaller networks in self-supervised scenarios. In this work, we concentrate on discriminative methods due to their benefits in network size and data efficiency.

Contrastive learning, a prevalent discriminative approach for self-supervised image learning, has been extensively developed \cite{Chen.05042021,Chen.17062020,Grill.14062020}. It strives to generate meaningful representations by distinguishing augmented image versions from distinct images. Deep learning models must comprehend and identify image semantics to achieve this, thus creating meaningful representations. \citet{Patacchiola.10062020} demonstrate that employing a classification head (relation module) on these representations reformulates the contrastive objective into a classification objective. This eliminates the need for typically large batch sizes and provides a strong supervisory signal, that is not prone to collapsing.

\begin{figure}[t]
\includegraphics[width=14cm]{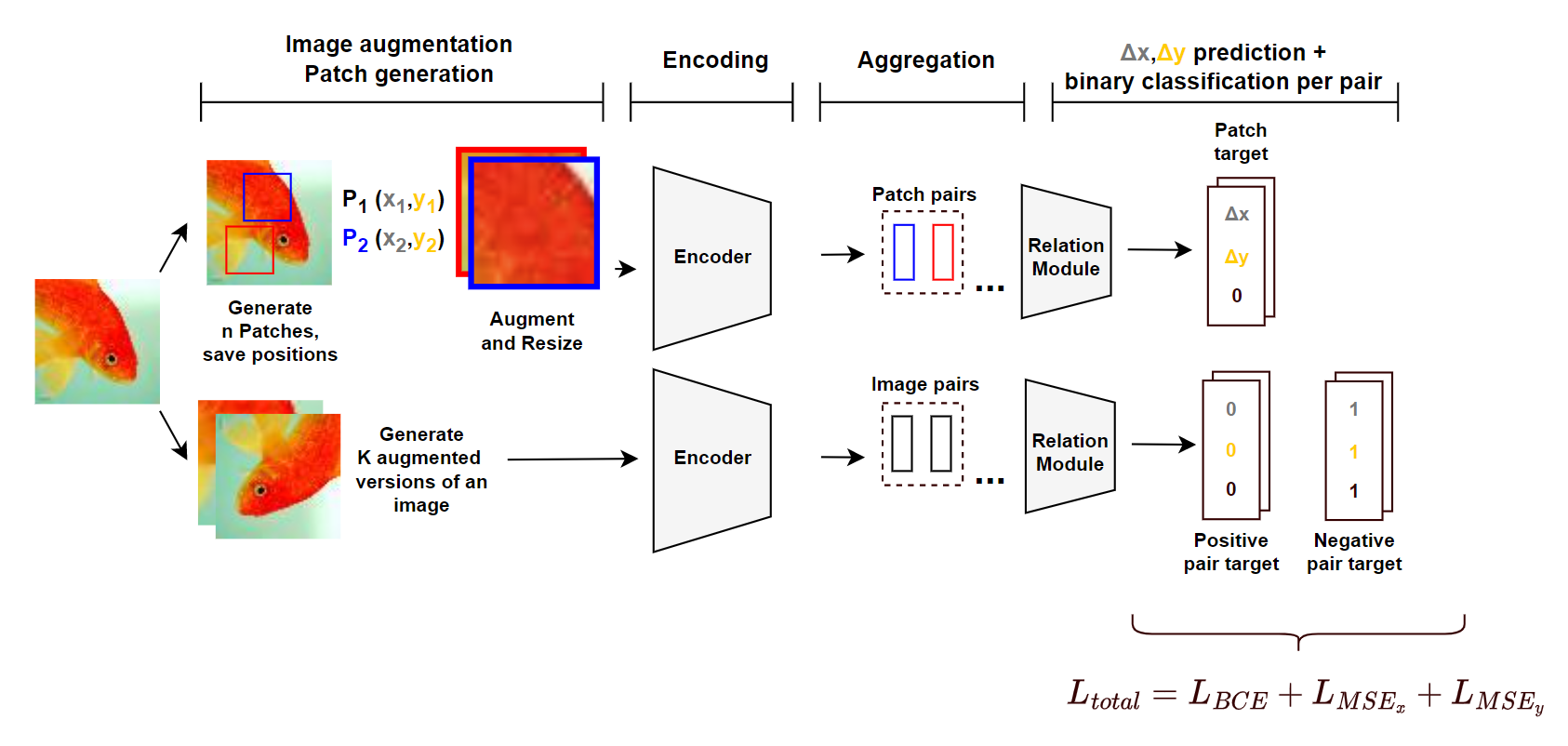}
\caption{The original augmentation process is expanded by the selection and rescaling of patches. After the prediction of the representations of both patches and images by the same encoder network, both types of representation are aggregated through specific methods into pairs. The patch aggregation is not using negative samples and defaults to a positive label for the contrastive objective. The image aggregation expands the target by pseudo labels for distance, depending on if the pair is positive or negative.}
\centering
\end{figure}

Leveraging the inherent capability of this formulation to flexibly accommodate supplementary objectives, wherein the classification head predicts the relative distance between two randomly selected patches from the same image. We argue that this objective forces the network to recognize the primary object and the spatial relationships among its constituting parts, enabling the creation of meaningful, scaled-domain representations. By concatenating patch-based and full-image representations, we have formulated an expanded representation that outperforms competitors in linear evaluation, even when training images and augmentations are limited. Although additional patch computations during the inference phase lead to increased computational requirements, we demonstrate that the number of patch representations at inference can be flexibly adjusted for more challenging tasks. Furthermore, we present an alternative formulation of spatial reasoning, called \textit{additive-patch-use}, that avoids the cost of additional computing at inference time. All code will be released on GitHub after publication.

In summary, our contributions can be summarised as follows:

\begin{itemize}
    \item We propose Spatial Reasoning and define a straightforward integration into Relational Reasoning architectures.
    \item We demonstrate better visual representations on several evaluation scenarios, with reduced training phase computation.
    \item We showcase performance for variable inference computing requirements.
    \item We present an alternative formulation of Spatial Reasoning, referred to as \textit{additive-patch-use}, circumventing increased inference computation, with minimal loss in performance.
    \item We provide recommendations on optimal patch sizes and the number of patches to be used during training.
\end{itemize}

\section{Related works}
There have been significant advances in the creation of meaningful representations through self-supervised pretraining. Self-supervised pretraining methods in the image domain can be divided into discriminative and generative approaches \cite{Chen.13022020}. With modern Vision Transformer (ViT) architectures, significant strides have been made in the generative direction. \citet{Atito.08042021} show state-of-the-art results in linear evaluation among generative methods. This is obtained by training the ViT model to reconstruct transformed versions of the same images. Transformations range from partial grey scale to random replacement of image regions with different images. Modern discriminative methods like MoCo v3 combined with ViT architectures lead to state-of-the-art results as well \cite{Chen.05042021}. Yet these architectures usually require high computational resources, especially during training. The ViT from  \citet{Atito.08042021}, for example, contains  86 Million parameters in its base configuration, while in \citet{Chen.05042021} variants with 300 million parameters and batch sizes up to 6000 are investigated. Additionally, training of ViT-based  models for generative tasks exhibits instabilities \cite{Chen.05042021}. There have been advances in training more data-efficient smaller transformers effectively, for example, reducing the size of ViT to 5 Million parameters in tiny variants, which allows training on 4 GPUs in less than three days on ImageNet\cite{Touvron.23122020}. Yet, pretraining transformer architectures seems not feasible for reaching good downstream performance with smaller datasets and fewer computational resources \cite{Zhai_2022_CVPR}. In contrast, Convolutional Neural Networks (CNNs), such as the Residual Network (ResNet) architectures, demonstrated competitive performance, even with pretraining on small datasets \cite{he2015deep}. Due to their design principles focusing on local receptive fields, shared weights, and spatial hierarchies, CNNs have a more restricted and efficient parameter space, which can be advantageous when training data is scarce \cite{LeCun.2015}. Specifically, ResNet architectures, which introduce "skip connections" or "shortcuts" to allow the gradient to be directly backpropagated to earlier layers, have shown impressive performance in various computer vision tasks \cite{he2015deep}. Therefore, convolutional architectures can serve as robust and effective alternatives to transformers in contexts where data or available compute is limited.

Prior research has explored the efficacy of incorporating spatial information contained in patches for enhancing representations in self-supervised learning. Specifically, \citet{Noroozi.30032016} use jigsaw puzzles and train neural networks to predict the correct combination to solve them. The proposed approach operates on nine patches using a Siamese-like architecture. 
Several other methods have been proposed for solving different variants of jigsaw puzzles. \citet{Kim.06022018} addressed the problem of solving a damaged jigsaw puzzle by reconstructing a missing piece. \citet{Wei.02122018} generalized the jigsaw problem and proposed an iterative solution. In contrast, our approach utilizes a network that has limited information about the image, with its relation module being fed with only two patches at a time. We argue that this limitation encourages the encoder to reason more deeply about the structure of the partially unseen object. Additionally, the network architecture in our approach is simpler compared to those in previous works and can be readily integrated into existing discriminatory approaches.

 A closer approach to our proposal is presented in \citet{Doersch.19052015}. Here the authors randomly sample a patch and then another at 8 adjacent locations. They then use a Siamese-like network to predict the relative positions as a probability over the eight possible patch locations. To avoid trivial solutions, the authors introduce a small location jitter as well as gaps between patches. In contrast, we simplify this approach by sampling at two completely random, non-overlapping positions, and formulate the problem as a contrastive objective. We additionally make use of the standard set of augmentations introduced by \citet{Patacchiola.10062020}, to combat other trivial solutions. 
 
 It is worth mentioning that all the approaches mentioned above generally require larger architectures with shared weights, making them challenging to be used in practice. We solve this problem by leveraging aggregation techniques \cite{Patacchiola.10062020}.

\section{Spatial reasoning}
In object recognition tasks, it may, for example, be enough to identify the existence of an elephant skin pattern to discriminate between specific pictures in a dataset. However, this local reasoning might result in poor generalization performance \cite{geirhos2018imagenet}. Traditional contrastive learning approaches use augmentations like cropping to combat this issue. With Spatial Reasoning, we aim to strengthen the supervisory signal and create more meaningful representations that not only contain the necessary information to identify an object but encode spatial relationships between its constituting parts, or between the main object and the background. As Spatial Reasoning builds on the idea that it is possible to create meaningful representation even from small patches of the image, we set up the objective of predicting the relative distance between N randomly sampled patches from the same image as an additional auxiliary prediction target. By using a classification head to reformulate typical contrastive losses as a classification problem, the introduction of Spatial Reasoning as auxiliary tasks is straightforward.

\subsection{Patch and label generation}
In standard Relational Reasoning \cite{Patacchiola.10062020}, an image is augmented K times, and each of the augmented versions is then provided to the network, combined with augmented versions of other images from the same batch. The actual batch size thus corresponds to $K$ times the number of images loaded per batch. We augment this process by creating N random positions per unaugmented image and extracting square patches at the chosen locations. The first two extracted patches are always guaranteed to be non-overlapping. This reduces the chance of finding a trivial solution. After extracting the patches, they are scaled to the standard input size and saved in the input batch with their corresponding locations as targets. To further reduce the chance of easy solutions, each patch is separately transformed by random colour jittering and grayscaling. We keep the augmentation scheme (colour jitter, grayscale, random resized crop, random horizontal flip) for the full-sized images, as introduced in Relational Reasoning.
Saving patches not directly as pairs allow the usage of the algorithm implemented in \cite{Patacchiola.10062020} for live pair aggregation with minimal adjustments.

After generating patches and locations, they are appended to the output of the standard image augmentation step. 
We additionally test a different version of our approach, in which 
selected patches are added in their original resolution onto a black image. This can be obtained by padding the patch back to the original size with zeros and is regarded to as \textit{additive-patch-use} in the following discussion. We discuss this approach in more detail in subsection \ref{sec:additivepatchuse} of the Experiment.

\subsection{Patch position prediction}
During training, the patches are fed into the encoder network with the augmented images. The total number of representations can be expressed as

\begin{equation}
P = K \times M + M \times N,
\label{eq:pairs}
\end{equation}

where $K$ is the number of augmentations, $N$ is the number of patches generated per image and $M$ is the size of the mini-batch. Afterward, an aggregation function is used, which concatenates two representations to be fed as input for the classification head. The number of generated pairs again depends on the mini-batch size $M$, as well as the number of augmentations $K$ and the number of patches $N$. The standard aggregation function creates all possible positive representation pairs by shifting the corresponding vectors. For each positive pair, we also consider a negative pair generated by shifting to the next image. The number of total combinations given by just the augmentations is given in \citet{Patacchiola.10062020} as 
$A = M(K^2 - K)$. We extend this number to 

\begin{equation}
A' = M(K^2 - K) + M((N^2 - N)/2),
\label{eq:extended_pairs_corrected}
\end{equation}

where $N$ is the number of patches generated per image, and $(N^2 - N)/2$ represents the number of unique patch pairs generated by iterating over the total number of patches. Note that the negative samples are not used for the patch pairs. Our tests show that negative patch pairs do not increase the downstream performance, as a negative patch pair would provide the same signal as a normal negative augmented representation pair from Relational Reasoning, just with less variance in augmentation.
Omitting the generation of negative pairs results in a reduced scaling for the total number of aggregated patch representations while not reducing performance. 

The relation module is adapted from \cite{Patacchiola.10062020} and expanded to 3 neurons in the final layer. The first neuron is used for the classification of negative and positive samples. The other two neurons predict the $x$- and $y$-coordinates of the relative distance of the patches or the dummy targets for standard images. In the case the source is an augmented image and not a patch, these units are used as pseudo targets for the classificaton. The idea is to set the target to zero if the image pair is positive and to one if the pair contains different images. With this adjustment, the additional two target neurons propagate gradients even for the full-sized augmented image pairs. The patches are aggregated in a similar way, with only minor differences. First, the shifting aggregation is set to the number of generated patches, and the positional labels are subtracted from each other to produce the relative distance. For example, if patch $p_1$ has a position of $(0.2, 0.6)$ and patch $p_2$ has a position of $(0.4, 0.1)$, the resulting target would be $p_1 - p_2 = (-0.2, 0.5)$.  

Let $L_{BCE}$ denote the binary cross-entropy loss for the first neuron, which classifies negative and positive samples, and let $L_{MSE_x}$ and $L_{MSE_y}$ represent the mean-squared-error losses for the second and third neurons, which predict the x- and y-coordinates of the relative distance of the patches or the dummy targets. Thus, the total loss $L_{total}$ can be defined as 

\begin{equation}
L_{total} = L_{BCE} + L_{MSE_x} + L_{MSE_y}.
\end{equation}

\subsection{Dynamic compute requirements in the evaluation procedure}
At inference time, we separate the image into $n$ patches of the same size used during training and concatenate their representations with the representation from the whole image. It may be noted that most of our experiments use nine patches that cover the whole image in total and lead to a small overlap. The actual impact on downstream performance is investigated in section \ref{sec:dynamiccompute}. Using $n$ patches leads to an $n$ times higher computational cost at inference. On the other hand, our method is mostly trained with two generated patches per image and K = 4 augmentations. While still outperforming the reported results with values for K ranging from 16 to 32 \cite{Patacchiola.10062020}, which leads to a significantly lower computation demand in training. 

In essence, for successful spatial reasoning, the neural network is required to generate representations from individual patches that empower the relation module to predict the relative distance between these patches accurately. Given the fact that our relation module is only comprised of two layers, the encoder network has a crucial role to play: it must effectively discern which part of an object corresponds to the given patch. This integral information, once encoded, is expected to enhance the quality of the final representation when combined with representations from other patches. In other words, by forcing the network to understand the spatial relationships within an image, we encourage the development of more robust and meaningful representations that go beyond simple object identification to include spatial context.

\section{Experiments}
In the process of finding a simple formulation for spatial reasoning, we tested several different configurations of Spatial Reasoning and the corresponding pretraining architecture. After some experimentation with stepwise distance prediction and negative patches, we chose the simple yet flexible formulation explained above. In this chapter, we will elaborate on the results of our experiments regarding the impact of patch size and number of patches on the performance in linear evaluation. Except for the concatenation of representations in the linear evaluation, we generally follow the procedure given in \cite{Patacchiola.10062020}. Specifically, we train the backbone model for 200 epochs using the unlabeled training dataset. Subsequently, a linear classifier is trained for 100 epochs, utilizing the features extracted from the backbone (without performing backpropagation on the backbone weights). The accuracy achieved by this classifier on the test dataset is regarded as the ultimate metric for evaluating the quality of the representations. We also do not change the augmentation strategy, except for leaving out cropping for patches. 

In Table \ref{tab:comparison}, we present a comparison of the performance of our method on various benchmarks. The table includes the mean accuracy (in percentage) and standard deviation over three runs using ResNet-32 for all datasets except STL-10, where ResNet-34 is used. The performance of our method is compared across different datasets, such as CIFAR-100, tiny-ImageNet, CIFAR-100-20 (coarse-grained), and STL-10. Additionally, we compare the performance on two cross-domain tasks: 10$\rightarrow$100 (training on CIFAR-10 and testing on CIFAR-100) and 100$\rightarrow$10 (training on CIFAR-100 and testing on CIFAR-10). The table highlights the effectiveness of our method in direct comparison to previous results in various settings. 
For tiny-imagenet, we reach a performance of 33.08\% with four augmentations compared to the 30.5\% reported in \cite{Patacchiola.10062020} with 16 augmentations on a ResNet-32. This is obtained using only two patches per image during training, and the optimal size of 24x24 pixels per patch, see figures \ref{fig:patchsizediagram} and \ref{fig:patchnumberdiagram}. In the remaining chapter, we provide insights into how the choice of patch size and number of patches affects the performance in linear evaluation.

\pgfplotstableread{
method value1 value2 value3
CIFAR-100      50.32 49.64 50.59
tiny-ImgNet    33.04 33.16 33.03
10->100        48.52 47.45 47.83
100->10        75.94 74.98 76.47
CIFAR-100-20   58.65 58.72 58.17
STL-10         90.24 89.71 90.80
}\ourdatatable
\setlength{\tabcolsep}{1pt}

\begin{table}[h]
\small
\centering
\caption{Comparison on various benchmarks. Mean accuracy (percentage) and standard deviation over three runs (ResNet-32/ResNet-34 for STL-10). Best results in bold.}
\label{tab:comparison}
\begin{tabular}{lcccccc}

\toprule
Method & CIFAR-100 & tiny-ImgNet & 10$\rightarrow$100 & 100$\rightarrow$10 & CIFAR-100-20 & STL-10 \\
\midrule
\makecell[l]{Supervised \\ (upper bound)} & \textbf{65.32$\pm$0.22} & \textbf{50.09$\pm$0.32} & \textbf{33.98$\pm$0.71} & \textbf{71.01$\pm$0.44} & \textbf{76.35$\pm$0.57} & \textbf{69.82$\pm$3.36} \\

\makecell[l]{Random Weights \\ (lower bound)} & 7.65$\pm$0.44 & 3.24$\pm$0.43 & 7.65$\pm$0.44 & 27.47$\pm$0.83 & 16.56$\pm$0.48 & n/a \\

\makecell[l]{DeepCluster \\ \cite{Caron.15072018}} & 20.44$\pm$0.80 & 11.64$\pm$0.21 & 18.37$\pm$0.41 & 43.39$\pm$1.84 & 29.49$\pm$1.36 & 73.37$\pm$0.55 \\

\makecell[l]{RotationNet \\ \cite{Gidaris.21032018}} & 29.02$\pm$0.18 & 14.73$\pm$0.48 & 27.02$\pm$0.20 & 52.22$\pm$0.70 & 40.45$\pm$0.39 & 83.29$\pm$0.44 \\

\makecell[l]{Deep InfoMax \\ \cite{Hjelm.20082018}} & 24.07$\pm$0.05 & 17.51$\pm$0.15 & 23.73$\pm$0.04 & 45.05$\pm$0.24 & 33.92$\pm$0.34 & 76.03$\pm$0.37 \\

\makecell[l]{SimCLR \\ \cite{Chen.13022020}} & 42.13$\pm$0.35 & 25.79$\pm$0.35 & 36.20$\pm$0.16 & 65.59$\pm$0.76 & 51.88$\pm$0.48 & 89.31$\pm$0.14 \\

\makecell[l]{Relational Reasoning \\ \cite{Patacchiola.10062020}} & 46.17$\pm$0.17 & 30.54$\pm$0.42 & 41.50$\pm$0.35 & 67.81$\pm$0.42 & 52.44$\pm$0.47 & 89.67$\pm$0.33 \\
\makecell[l]{\textbf{Ours}}& 
\multicolumn{1}{c}{\pgfplotstablegetelem{0}{value1}\of{\ourdatatable}\pgfmathsetmacro{\valA}{\pgfplotsretval}\pgfplotstablegetelem{0}{value2}\of{\ourdatatable}\pgfmathsetmacro{\valB}{\pgfplotsretval}\pgfplotstablegetelem{0}{value3}\of{\ourdatatable}\pgfmathsetmacro{\valC}{\pgfplotsretval}\pgfmathsetmacro{\cifaravg}{(\valA + \valB + \valC)/3}\pgfmathsetmacro{\cifarstd}{sqrt((pow(\valA-\cifaravg,2) + pow(\valB-\cifaravg,2) + pow(\valC-\cifaravg,2))/2)}\textbf{\pgfmathprintnumber[assume math mode=true]{\cifaravg}$\pm$\pgfmathprintnumber[assume math mode=true]{\cifarstd}}}&
\multicolumn{1}{c}{\pgfkeys{/pgf/number format/.cd,fixed,precision=2}\pgfplotstablegetelem{1}{value1}\of{\ourdatatable}\pgfmathsetmacro{\valA}{\pgfplotsretval}\pgfplotstablegetelem{1}{value2}\of{\ourdatatable}\pgfmathsetmacro{\valB}{\pgfplotsretval}\pgfplotstablegetelem{1}{value3}\of{\ourdatatable}\pgfmathsetmacro{\valC}{\pgfplotsretval}\pgfmathsetmacro{\tinyavg}{(\valA + \valB + \valC)/3}\pgfmathsetmacro{\tinystd}{sqrt((pow(\valA-\tinyavg,2) + pow(\valB-\tinyavg,2) + pow(\valC-\tinyavg,2))/2)}\textbf{
\pgfmathprintnumber[assume math mode=true]{\tinyavg}$\pm$\pgfmathprintnumber[assume math mode=true]{\tinystd}}}&
\pgfplotstablegetelem{2}{value1}\of{\ourdatatable}\pgfmathsetmacro{\valA}{\pgfplotsretval}\pgfplotstablegetelem{2}{value2}\of{\ourdatatable}\pgfmathsetmacro{\valB}{\pgfplotsretval}\pgfplotstablegetelem{2}{value3}\of{\ourdatatable}\pgfmathsetmacro{\valC}{\pgfplotsretval}\pgfmathsetmacro{\tenavg}{(\valA + \valB + \valC)/3}\pgfmathsetmacro{\tenstd}{sqrt((pow(\valA-\tenavg,2) + pow(\valB-\tenavg,2) + pow(\valC-\tenavg,2))/2)}\textbf{\pgfmathprintnumber[assume math mode=true]{\tenavg}$\pm$\pgfmathprintnumber[assume math mode=true]{\tenstd}}&
\pgfplotstablegetelem{3}{value1}\of{\ourdatatable}\pgfmathsetmacro{\valA}{\pgfplotsretval}\pgfplotstablegetelem{3}{value2}\of{\ourdatatable}\pgfmathsetmacro{\valB}{\pgfplotsretval}\pgfplotstablegetelem{3}{value3}\of{\ourdatatable}\pgfmathsetmacro{\valC}{\pgfplotsretval}\pgfmathsetmacro{\hundredavg}{(\valA + \valB + \valC)/3}\pgfmathsetmacro{\hundredstd}{sqrt((pow(\valA-\hundredavg,2) + pow(\valB-\hundredavg,2) + pow(\valC-\hundredavg,2))/2)}\textbf{\pgfmathprintnumber[assume math mode=true]{\hundredavg}$\pm$\pgfmathprintnumber[assume math mode=true]{\hundredstd}}&
\pgfplotstablegetelem{4}{value1}\of{\ourdatatable}\pgfmathsetmacro{\valA}{\pgfplotsretval}\pgfplotstablegetelem{4}{value2}\of{\ourdatatable}\pgfmathsetmacro{\valB}{\pgfplotsretval}\pgfplotstablegetelem{4}{value3}\of{\ourdatatable}\pgfmathsetmacro{\valC}{\pgfplotsretval}\pgfmathsetmacro{\hundredavg}{(\valA + \valB + \valC)/3}\pgfmathsetmacro{\hundredstd}{sqrt((pow(\valA-\hundredavg,2) + pow(\valB-\hundredavg,2) + pow(\valC-\hundredavg,2))/2)}\textbf{\pgfmathprintnumber[assume math mode=true]{\hundredavg}$\pm$\pgfmathprintnumber[assume math mode=true]{\hundredstd}}&
\pgfplotstablegetelem{5}{value1}\of{\ourdatatable}\pgfmathsetmacro{\valA}{\pgfplotsretval}\pgfplotstablegetelem{5}{value2}\of{\ourdatatable}\pgfmathsetmacro{\valB}{\pgfplotsretval}\pgfplotstablegetelem{5}{value3}\of{\ourdatatable}\pgfmathsetmacro{\valC}{\pgfplotsretval}\pgfmathsetmacro{\hundredavg}{(\valA + \valB + \valC)/3}\pgfmathsetmacro{\hundredstd}{sqrt((pow(\valA-\hundredavg,2) + pow(\valB-\hundredavg,2) + pow(\valC-\hundredavg,2))/2)}\textbf{\pgfmathprintnumber[assume math mode=true]{\hundredavg}$\pm$\pgfmathprintnumber[assume math mode=true]{\hundredstd}}
\\
\hline
\end{tabular}
\end{table}

\subsection{Patch size}
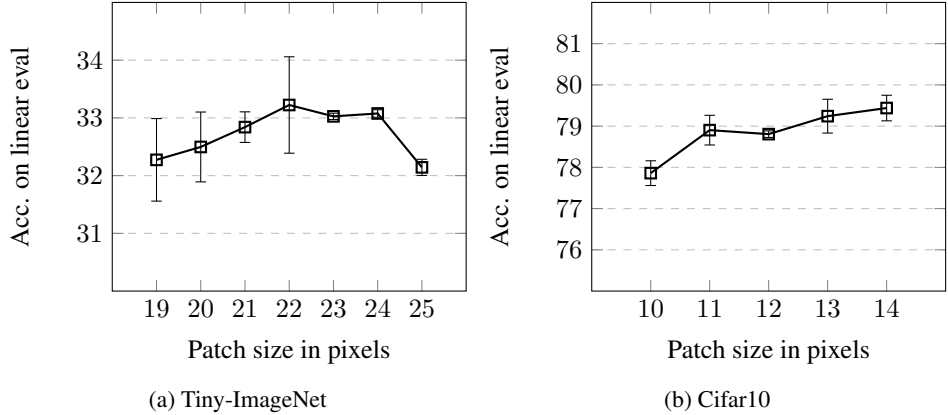
\begin{figure}[h]
\centering
\begin{subfigure}{.45\textwidth}
\begin{tikzpicture}
\begin{axis}[
    width=\linewidth,
    xlabel={Patch size in pixels},
    ylabel={Acc. on linear eval},
    xmin=18, xmax=26,
    ymin=30, ymax=35,
    xtick={19,20,21,22,23,24,25},
    ytick={31,32,33,34},
    legend pos=north east,
    ymajorgrids=true,
    grid style=dashed,
    error bars/y dir=both,
    error bars/y explicit,
]

\addplot[
    color=black,
    mark=square,
    thick,
    error bars/.cd,
    y dir=both,
    y explicit,
    ]
    table [
    x=patchsize,
    y expr= (\thisrow{value1} + \thisrow{value2} + \thisrow{value3})/3,
    y error expr= {abs((max(\thisrow{value1}, \thisrow{value2}, \thisrow{value3}) - min(\thisrow{value1}, \thisrow{value2}, \thisrow{value3}))/2)}
    ] {\datatable};
    
\end{axis}
\end{tikzpicture}
\caption{Tiny-ImageNet}
\label{fig:tinyimagenet}
\end{subfigure}
\begin{subfigure}{.45\textwidth}
\begin{tikzpicture}
\begin{axis}[
    width=\linewidth,
    xlabel={Patch size in pixels},
    ylabel={Acc. on linear eval},
    xmin=9, xmax=15,
    ymin=75, ymax=82,
    xtick={10,11,12,13,14},
    ytick={76,77,78,79,80,81},
    legend pos=north east,
    ymajorgrids=true,
    grid style=dashed,
    error bars/y dir=both,
    error bars/y explicit,
]

\addplot[
    color=black,
    mark=square,
    thick,
    error bars/.cd,
    y dir=both,
    y explicit,
    ]
    table [
    x=patchsize,
    y expr= (\thisrow{A1} + \thisrow{A2} + \thisrow{A3})/3,
    y error expr= {abs((max(\thisrow{A1}, \thisrow{A2}, \thisrow{A3}) - min(\thisrow{A1}, \thisrow{A2}, \thisrow{A3}))/2)}
    ] {\datatableCifar};
    
\end{axis}
\end{tikzpicture}
\caption{Cifar10}
\label{fig:cifar10}
\end{subfigure}
\caption{Impact of patch size on the linear evaluation performance on Tiny-ImageNet and Cifar10. Standard evaluation procedure with 100 epochs of finetuning on ten combined representations(1 image + 10 patches) and 2 patches extracted per image for each epoch.}
\label{fig:patchsizediagram}
\end{figure}

As evidenced by previous studies \cite{Krizhevsky.2012, Chen.13022020}, the application of cropping and zooming as augmentation step significantly influences the performance of subsequent processes. This observation extends to our own experiments concerning the size of sampled patches, as illustrated in Figure \ref{fig:patchsizediagram}. The best performance for tiny-imagenet is reached for 23 and 24 pixels in width and length. Lower performance with smaller patches is most likely due to the missing amount of relevant information from the main object in the selected patch. This might lead the network to learn more abstract models based on the background and not on the object that is to be identified, which is an instance of the shortcut learning problem \cite{Geirhos.2020}. The lower performance with larger patches is likely due to lower task difficulty. In fact, if the patches are too large, the number of possible non-overlapping locations decreases substantially. This leads to an easier task that reduces the strength of the supervisory signal produced by Spatial Reasoning.

\subsection{Number of patches}
\begin{figure}[t]
\begin{tikzpicture}
\begin{axis}[
    xlabel={Patch number per base image},
    ylabel={Accuracy on linear evaluation},
    xmin=1.5, xmax=6.5,
    ymin=32, ymax=36,
    xtick={2,3,4,5,6},
    ytick={32,33,34,35,36},
    legend pos=north east,
    ymajorgrids=true,
    grid style=dashed,
    width=0.4\textwidth,  
    height=0.4\textwidth, 
]

\addplot[
    color=black,
    mark=square,
    ]
    coordinates {
    (2,33.04)(3,33.20)(4,33.88)(5,33.44)(6,33.01)
    };
    \legend{ResNet32}
    
\end{axis}
\end{tikzpicture}
\centering
\caption{Impact of the number of patches extracted from each image on the linear evaluation performance on tiny-imagenet. Standard evaluation procedure with 100 epochs of finetuning on 10 combined representations(1 image + 9 patches) and 4 augmentations.}
\label{fig:patchnumberdiagram}
\centering
\end{figure}
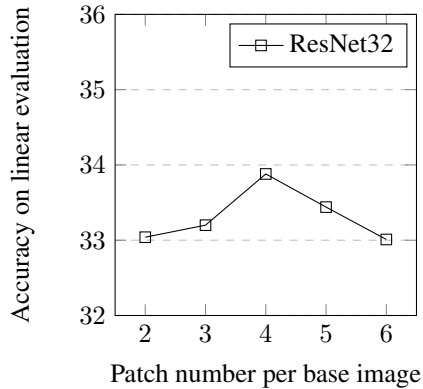

This section discusses the impact of the number of patches extracted from an image in the training procedure on the linear evaluation performance. The results seen in figure \ref{fig:patchnumberdiagram} must be evaluated in combination with the number K of augmentations. As we use K=4 in all of our experiments, a patch number of 2 means that every third input image is a patch. This also directly influences the number of possible combinations that are aggregated from the representations. As can be seen from figure \ref{fig:patchnumberdiagram}, the optimal number of patches is 3, which means 3/7 of the input images are patches, and 192 of 960 total aggregated representation pairs are patch based in an optimal setting.

The lower performance with a rising number of patches can  be explained by the higher impact of the \textit{zoomed-in} domain. If too many images differ from the domain of the original dataset, the performance decreases due to the domain generalization problem \cite{Torralba.2011, ganin2015unsupervised}. As roughly half of all created representation pairs are from patches with a patch count of 6, the performance decays to 33\%. The initial improvement by 1\% from two to four patches per base image proves the positive impact of spatial reasoning.

\pgfplotstableread{
patchsize value1 value2 value3
19        31.47  32.45  32.9
20        31.97  33.18  32.34
21        32.5  33.03  32.99
22        32.35  33.3   34.02
23        32.97  33.03  33.08
24        33.04  33.16  33.03
25        32.02  32.11  32.3
}\datatable

\pgfplotstableread{
patchsize A1      A2      A3
10        78.23   77.72   77.63
11        79.16   78.44   79.11
12        78.75   78.91   78.76
13        79.69   78.87   79.17
14        79.50   79.10   79.72
}\datatableCifar

\subsection{Impact of dynamic compute}\label{sec:dynamiccompute}
Prior discussions have underscored the role played by the number of image patches extracted during the training process and its subsequent effect on linear evaluation performance. An integral part of this evaluation procedure involves the concatenation of image representations with their corresponding patch representations. While this process necessitates an increased computational load during the inference phase, it concurrently enables the incorporation of dynamic computing at inference.

\begin{figure}[ht]
\begin{tikzpicture}
\begin{axis}[
    xlabel={Number of patches},
    ylabel={Acc. on linear eval},
    xmin=0, xmax=10,
    ymin=30, ymax=34.5,
    xtick={1,3,5,7,9},
    ytick={30.5,31,31.5,32,32.5,33,33.5,34},
    legend pos=south east,
    ymajorgrids=true,
    grid style=dashed,
    error bars/y dir=both,
    error bars/y explicit,
    width=0.7\textwidth,  
    height=0.4\textwidth, 
    legend entries={Augmented, Unaugmented},
]

\addplot[
    color=red,
    mark=square,
    thick,
    error bars/.cd,
    y dir=both,
    y explicit,
    ]
    table [
    x=patchsize,
    y expr= (\thisrow{A1} + \thisrow{A2} + \thisrow{A3})/3,
    y error expr= {abs((max(\thisrow{A1}, \thisrow{A2}, \thisrow{A3}) - min(\thisrow{A1}, \thisrow{A2}, \thisrow{A3}))/2)}
    ] {\datatabledynamicaugment};  
\addplot[
    color=black,
    mark=square,
    thick,
    error bars/.cd,
    y dir=both,
    y explicit,
    ]
    table [
    x=patchsize,
    y expr= (\thisrow{A1} + \thisrow{A2} + \thisrow{A3})/3,
    y error expr= {abs((max(\thisrow{A1}, \thisrow{A2}, \thisrow{A3}) - min(\thisrow{A1}, \thisrow{A2}, \thisrow{A3}))/2)}
    ] {\datatabledynamic};
\end{axis}
\end{tikzpicture}
\centering
\caption{Comparative Influence of Dynamic Compute with and without Augmentations. This figure illustrates the impact of varying patch numbers on the accuracy of linear evaluation, represented in black for unaugmented overfitting and in red for augmented configurations. The error bars represent the range of accuracy values obtained from three independent experiments. It is discernible that the incorporation of additional augmentations in the red graph aids in mitigating overfitting and enhancing accuracy, particularly with larger representations.}
\label{fig:dynamiccompute}
\end{figure}
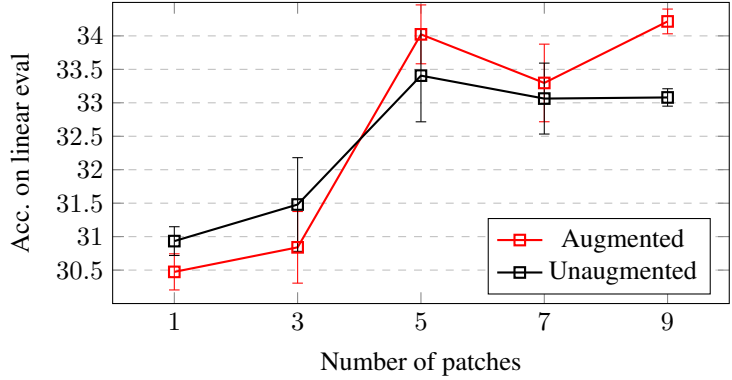

Figure \ref{fig:dynamiccompute} demonstrates the impact of varying patch numbers at the inference stage on the accuracy yielded in linear evaluation. This experiment was conducted using two patches and a constant number of augmentations $K=4$ during training.

Here, we designate the central patch as the first one, while the patches horizontally adjacent are referred to as patches 2 and 3, respectively. Patches 5 and 6 correspond to the patches situated at the mid-bottom and mid-top positions. When employing five patches, we sample from positions 1 to 5, and in the case of seven patches, two additional random corner patches are sampled.

The initial increase in accuracy to 33.5\%  until the fifth patch's introduction can be attributed to the incremental availability of more information. However, the subsequent plateauing of performance upon increasing the patch number can be ascribed to the detrimental effects of overfitting. As the standard linear evaluation process excludes the use of any augmentations, the extensive size of the resultant representation over the 100 epochs in the evaluation procedure leads to a training accuracy of 45\%, as opposed to a test set accuracy of 33\%.

To substantiate this hypothesis, we also present results incorporating additional affine transformations in figure \ref{fig:dynamiccompute}, configured analogously to those in STL-10, barring flipping and cropping, on the training set, indicated in red. The results show a distinct improvement of up to 1.5\% with the use of nine patches. This underscores the potential for mitigating overfitting and fostering enhanced generalization by incorporating additional augmentations alongside larger representations.
\pgfplotstableread{
patchsize A1      A2      A3
1         31.16   30.91   30.73
3         32.32   30.92   31.20
5         33.22   34.19   32.81
7         32.58   33.64   32.97
9         33.23   33.04   32.97
}\datatabledynamic
\pgfplotstableread{
patchsize A1      A2      A3
1         30.4   30.24   30.78
3         30.93   31.33   30.26
5         34.54   33.66   33.87
7         34.06   32.93   32.9
9         34.44   34.14   34.07
}\datatabledynamicaugment

\subsection{Additive-patch-use} \label{sec:additivepatchuse}
The experiments and evaluations reveal certain limitations associated with spatial reasoning. Specifically, the training process requires a limit on the number of patches to mitigate the impact of domain shift as the number of patches increases. In addition, during evaluation and finetuning, there is a higher computational demand compared to similar methods because the patch representations have to be encoded. 

To combat these issues, it is necessary to eliminate the impact of domain shift and make it possible for the network to predict a single representation right away. To reach this goal, we experimented with an adapted version of spatial reasoning. In this approach, instead of scaling the extracted patches, we pad the extracted patch to the size of the base image. We refer to this approach as \textit{additive-patch-use}. The new images can be used for training instead of the resized patch. Additive-patch-use should reduce the domain shift, as the resolution of the patch itself is not changed and the network is able to extract the same features in the same resolution as in the normal base images. This makes the network training more robust to a higher number of patches. After training, we are now able to process an image with a single forward pass. This obviously limits the amount of information that can be extracted from each patch into 64 dimensions but should still take advantage of the deeper understanding of the object. 

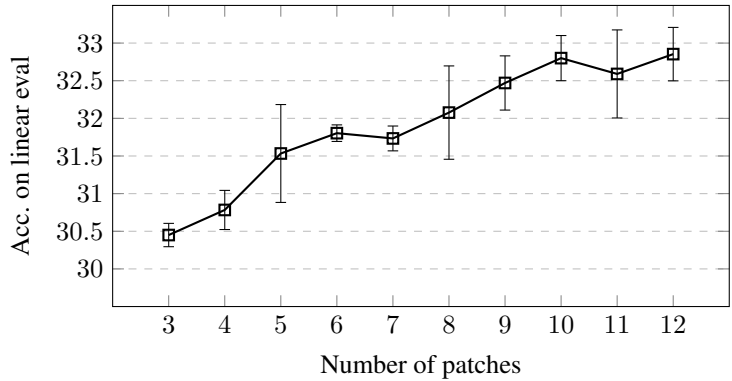
\begin{figure}[ht]
\begin{tikzpicture}
\begin{axis}[
    xlabel={Number of patches},
    ylabel={Acc. on linear eval},
    xmin=2, xmax=13,
    ymin=29.5, ymax=33.5,
    xtick={3,4,5,6,7,8,9,10,11,12},
    ytick={30,30.5,31,31.5,32,32.5,33},
    legend pos=north east,
    ymajorgrids=true,
    grid style=dashed,
    error bars/y dir=both,
    error bars/y explicit,
    width=0.7\textwidth,  
    height=0.4\textwidth, 
]

\addplot[
    color=black,
    mark=square,
    thick,
    error bars/.cd,
    y dir=both,
    y explicit,
    ]
    table [
    x=patchsize,
    y expr= (\thisrow{A1} + \thisrow{A2} + \thisrow{A3})/3,
    y error expr= {abs((max(\thisrow{A1}, \thisrow{A2}, \thisrow{A3}) - min(\thisrow{A1}, \thisrow{A2}, \thisrow{A3}))/2)}
    ] {\datatable};
    
\end{axis}
\end{tikzpicture}
\centering
\caption{Linear evaluation of additive patch creation. Evaluated with a single representation on tiny-imageNet with K=4, 3 runs with different seeds for each.}
\label{fig:additivepatchtest}

\end{figure}

As shown in figure \ref{fig:additivepatchtest}, additive-patch-use with 12 generated patches per image results in a 2.3\% performance increase in linear evaluation compared to  \cite{Patacchiola.10062020}. This is achieved with the same number of images as input in training, as K still equals 4. The positive impact of the method slightly diminishes with a higher amount of patches. This can be intuitively explained, as more patches lead to more overlap, which could enable trivial solutions. This, in turn, weakens the supervisory signal. In a broader context, the incorporation of additive-patch-use appears to be a straightforward augmentation technique for enhancing representations. Although it exhibits inferior performance compared to the Spatial Reasoning method described earlier, it mitigates the associated increase in inference cost. Additionally, the additive approach demonstrates robustness in relation to hyperparameter selection, particularly the number of patches. In the previously described method of Spatial Reasoning, selecting a significantly larger number of patch pairs compared to the number of traditional representation pairs can lead to a decrease in overall performance. In contrast, the additive-patch-use  mitigates this issue by being less sensitive to the choice of this hyperparameter, allowing for a more reliable integration without negatively affecting performance.

\pgfplotstableread{
patchsize A1      A2      A3
3         30.64   30.33   30.38
4         30.91   30.98   30.46
5         30.96   32.26   31.38
6         31.88   31.87   31.66
7         31.89   31.75   31.56
8         32.68   32.11   31.44
9         32.81   32.51   32.09
10        32.42   32.96   33.02
11        32.02   32.56   33.19
12        33.12   33.03   32.41
}\datatable

\section{Discussion and conclusion}

Our work shows that a relation head can be used to design the learning of Spatial Reasoning, formulating it as an auxiliary pretraining objective. Our approach creates better visual representations, even while reducing the necessary compute in training. Based on the results of the linear evaluation, Spatial Reasoning improves the resulting representations significantly in all evaluation scenarios.  Further research in this direction could improve the performance of discriminative self-supervised pretraining and reduce the performance gap to generative methods with significantly smaller networks.

Despite reduced computational requirements during training, compared to similar work, our approach requires higher compute cost at inference time, as well as the careful setting of the number of patches as a hyperparameter. We presented an alternative formulation, named additive-patch-use, that demonstrates to be a good solution for these limitations. In fact, it reduces the impact of domain shift as the number of patches increases, and it eliminates the additional computational demand in encoding all patches during inference. 

Future work could explore the definition of sampling techniques for the patch size, instead of choosing a constant value. Furthermore, as our approach can easily be adapted to other frameworks, a promising direction is to incorporate Spatial Reasoning into standard contrastive frameworks, like SimCLR \cite{Chen.13022020}.

\bibliographystyle{apalike} 
\bibliography{newbibliography}

\section{Additional information}
In this section, we present further information on our experimental setup, as well as the used datasets.
\subsection{Experimental setup}
All models are trained and evaluated on one of two different nodes. The first node contains a single RTX3080 GPU and a modern 8-core CPU. This node is used for all experiments with a patch count smaller than 3 and every dataset except STL-10. The second node contains an A100 GPU and a similar processor. All models fit on their respective GPUs. The training duration and learning rate as well as the augmentation scheme for all datasets are equivalent to the ones proposed by \citet{Patacchiola.10062020}. The only differences in augmentations are the missing random resized crop and random horizontal flip for each patch. The only time the general procedure of linear evaluation is changed by adding affine augmentations refers to the experiments done for Figure \ref{fig:dynamiccompute} and is denoted in the corresponding chapter. If error bars are present in a figure, each value is evaluated on three different seeds, with the upper limit of the bar denoting the maximum value and the lower limit the minimum value. The marker marks the average of all three runs.
\subsection{Datasets}
To evaluate our method, we used the following datasets: CIFAR-100, CIFAR-100-20, CIFAR-100$\rightarrow$10, CIFAR-10$\rightarrow$100, tiny-ImageNet, and STL-10. Below, we provide a brief description of each dataset and their characteristics.

\begin{itemize}
\item \textbf{CIFAR-100}: The CIFAR-100 dataset \cite{Krizhevsky.2012} consists of 60,000 32x32 color images, divided into 100 classes with 600 images per class. The dataset is split into a training set of 50,000 images and a test set of 10,000 images. The images in CIFAR-100 are low-resolution, making it a challenging dataset for object recognition tasks.
\item \textbf{CIFAR-10}: The CIFAR-10 dataset \cite{Krizhevsky.2012} has the same structure of 50000 training images and 10000 testing images, but only contains 10 classes.
\item \textbf{CIFAR-100-20}: The CIFAR-100-20 is CIFAR 100 with course-grained classes, containing only 20 superclasses from the original 100. This set classes is used to evaluate the course-grained performance of the model on a smaller and less diverse dataset.

\item \textbf{CIFAR-100$\rightarrow$10}: In this cross-domain task, the model is trained on the CIFAR-100 dataset and tested on the CIFAR-10 dataset. This task evaluates the model's ability to transfer knowledge from a more complex and diverse dataset to a simpler one.

\item \textbf{CIFAR-10$\rightarrow$100}: This cross-domain task is the opposite of the previous one. The model is trained on the CIFAR-10 dataset and tested on the CIFAR-100 dataset. This task assesses the model's capacity to generalize and adapt its learned features from a simpler dataset to a more complex and diverse one.

\item \textbf{Tiny-ImageNet}: The tiny-ImageNet dataset is a downscaled version of the ImageNet dataset \cite{YaLe.2015}, consisting of 200 classes, each with 500 training images, 50 validation images, and 50 test images. The images are 64x64 pixels in size. Due to the small size and diversity of the images, tiny-ImageNet presents a challenging benchmark for image recognition algorithms.

\item \textbf{STL-10}: The STL-10 dataset \cite{Coates.2011} is an image recognition dataset containing 10 classes with 1,300 96x96 color images per class. The dataset is designed for unsupervised and supervised learning, with a training set of 5,000 labeled images, 100,000 unlabeled images, and 8,000 test images. The higher resolution and the presence of both labeled and unlabeled data make STL-10 a suitable dataset for evaluating the performance of self-supervised learning methods.
\end{itemize}

These datasets provide a diverse set of challenges for our method, allowing us to evaluate its performance and effectiveness across various contexts and tasks.


\end{document}